\documentclass[10pt,twocolumn,letterpaper]{article}

\usepackage[pagenumbers]{wacv} % To force page numbers, e.g. for an arXiv version

\definecolor{wacvblue}{rgb}{0.21,0.49,0.74}
\usepackage[pagebackref,breaklinks,colorlinks,allcolors=wacvblue]{hyperref}

\usepackage{graphicx}
\usepackage{cuted}
\usepackage{caption}
\usepackage[T1]{fontenc}      % gives access to \k and many other accents
\usepackage[utf8]{inputenc}   % pdflatex: usually default today, but harmless/safe

\def\wacvPaperID{442} % *** Enter the WACV Paper ID here
\def\confName{WACV}
\def\confYear{2026}

\title{Talking Head Generation via AU-Guided Landmark Prediction}
\author{
  Shao-Yu Chang \textsuperscript{1} \quad
  Jingyi Xu\textsuperscript{1} \quad
  Hieu Le\textsuperscript{2, 3} \quad
  Dimitris Samaras\textsuperscript{1} \\[1ex]
  \textsuperscript{1}Stony Brook University \quad
  \textsuperscript{2}EPFL \quad
  \textsuperscript{3}University of North Carolina at Charlotte
}

\begin{document}
\maketitle

\begin{strip}
  \centering
  \includegraphics[width=0.95\textwidth]{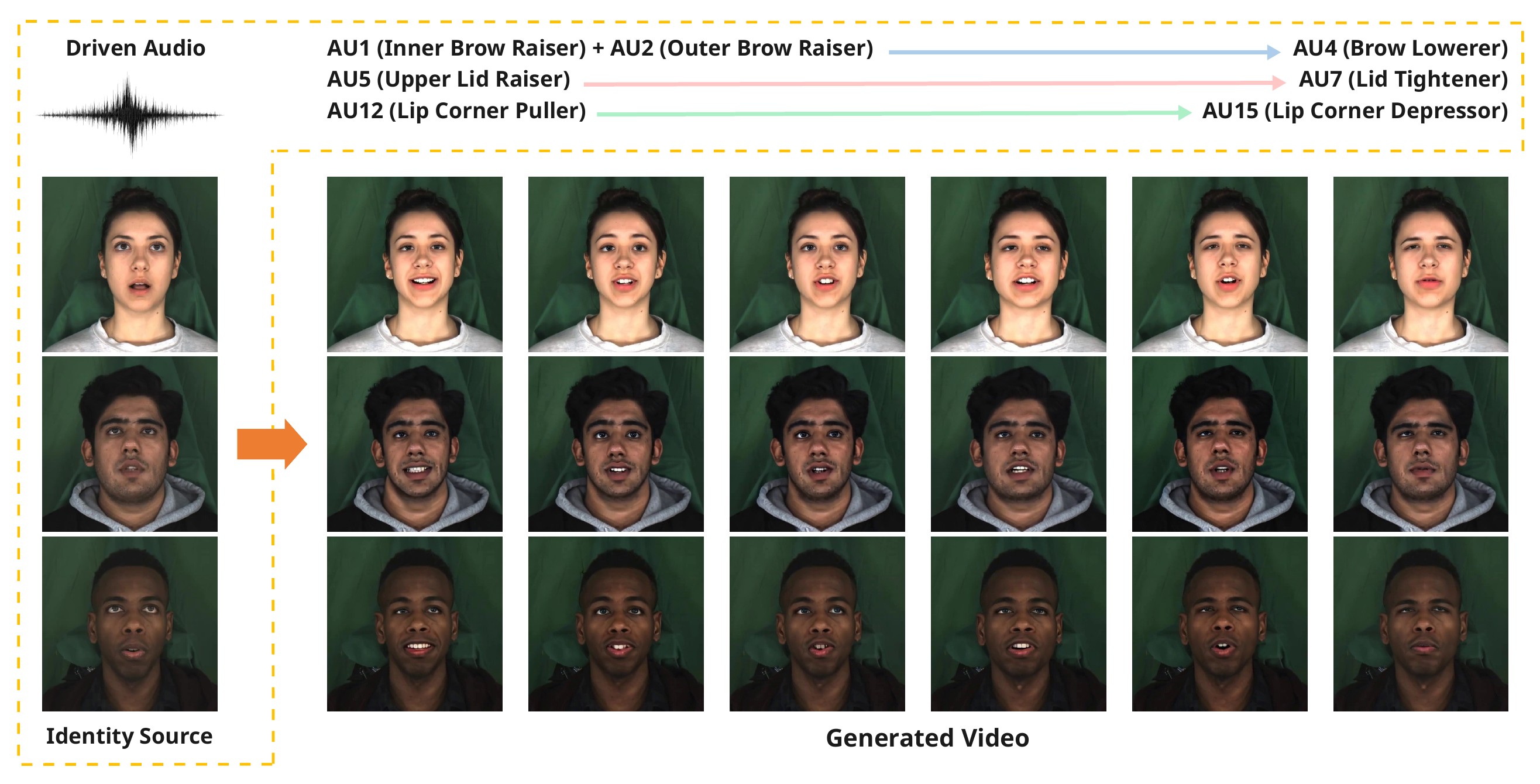}
  \captionof{figure}{\textbf{Illustrative results produced by our pipeline.} Given a single identity image, an audio clip, and per-frame AU intensity embeddings, our model generates a talking head whose expressions follow the specified muscle activations controlled by the driving AUs.}
  \label{fig:example}
\end{strip}
% TODO: Mention this figure somewhere in Introduction.
% In this example, we sequentially drive Inner Brow Raiser (AU 1)+Outer Brow Raiser (AU 2) → Brow Lowerer (AU 4), Upper Lid Raiser (AU 5) → Lid Tightener (AU 7), and Lip Corner Puller (AU 12) → Lip Corner Depressor (AU 15), producing the corresponding changes in eyebrow height, eyelid openness, and lip curvature.

\begin{abstract}
We propose a two-stage framework for audio-driven talking head generation with fine-grained expression control via facial Action Units (AUs). Unlike prior methods relying on emotion labels or implicit AU conditioning, our model explicitly maps AUs to 2D facial landmarks, enabling physically grounded, per-frame expression control. In the first stage, a variational motion generator predicts temporally coherent landmark sequences from audio and AU intensities. In the second stage, a diffusion-based synthesizer generates realistic, lip-synced videos conditioned on these landmarks and a reference image. This separation of motion and appearance improves expression accuracy, temporal stability, and visual realism. Experiments on the MEAD dataset show that our method outperforms state-of-the-art baselines across multiple metrics, demonstrating the effectiveness of explicit AU-to-landmark modeling for expressive talking head generation.
\end{abstract}
    
\section{Introduction}
\label{sec:intro}
Human faces are incredibly expressive. We communicate not only through words but also through subtle, dynamic facial cues such as a quick eyebrow raise, a small lip movement, or a brief blink. These micro-expressions convey rich emotional and social information and are central to how we signal intent, mood, and reaction in conversation. 

Replicating this level of subtlety in synthetic facial animation, however, remains a major challenge. While recent progress in audio-driven talking head generation has enabled the creation of realistic, lip-synced videos from speech, capturing the emotional nuance and speaker-specific variability found in natural human expression is still difficult. Many existing methods rely on discrete emotion labels~\cite{10.1145/3528233.3530745, Gan_2023_ICCV, 10658061, 10.1109/TMM.2021.3099900, 10.1145/3072959.3073658, 10049691, 10458318, Sinha2022EmotionControllableGT, inproceedings} or simply transfer expressions from a driving video~\cite{tan2024style2talker, cui2025hallo, wang2025emotivetalk}. While these approaches are effective to some extent, they often lack the granularity and interpretability needed for fine-grained emotional control, limiting their ability to model the continuous and nuanced activation of facial muscles.
%Talking head video generation has gained significant attention due to its applications in virtual communication, digital avatars, and content creation. Although recent approaches can generate visually realistic videos, they often rely on a small set of coarse emotion labels~\cite{10.1145/3528233.3530745, Gan_2023_ICCV, 10658061, 10.1109/TMM.2021.3099900, 10.1145/3072959.3073658, 10049691, 10458318, Sinha2022EmotionControllableGT, inproceedings} or simply copy expressions from driving videos~\cite{tan2024style2talker, cui2025hallo, wang2025emotivetalk}. Such methods cannot capture the subtle, person‑specific ways that the same emotion may manifest.
%For example, one speaker might widen their eyes to convey ``fear'', while another might lower their brows and partially close their eyelids for the same emotion. This limited granularity prevents modeling the continuous and nuanced activation of facial muscles.

%they often lack fine-grained control over facial expressions — particularly those that convey subtle and dynamic human emotions. Many existing methods depend on coarse emotion labels~\cite{10.1145/3528233.3530745, Gan_2023_ICCV, 10658061, 10.1109/TMM.2021.3099900, 10.1145/3072959.3073658, 10049691, 10458318, Sinha2022EmotionControllableGT, inproceedings} or directly transfer expressions from driving videos~\cite{tan2024style2talker, cui2025hallo, wang2025emotivetalk}, limiting their ability to model the continuous and nuanced activation of facial muscles.

Facial Action Units (AUs) offer a promising path toward more expressive control, as they provide a physically grounded representation of individual muscle group activations. For instance, FG-EmoTalk~\cite{FGEmoTalk} incorporates AUs as latent conditioning signals to guide expression synthesis. However, such implicit conditioning requires the network to learn a complex mapping from high-level AU intensities directly to low-level pixel geometry. Without explicit structural constraints, this often results in unnatural or unstable expressions — especially when modeling subtle muscle movements that demand precise geometric alignment.

%In this paper, we address this gap by introducing an explicit intermediate representation: instead of relying on the network to hallucinate geometry from AUs, we first predict a dynamic sequence of facial landmarks conditioned on the input AUs. This sequence acts as a spatial and temporal scaffold, capturing how muscle activations translate into concrete facial motion. We then use a diffusion model to synthesize the final frames, ensuring high-quality, temporally coherent talking head videos that reflect the intended expressions.

In this paper, we tackle this challenge by introducing a landmark-based intermediate representation that enables fine-grained and physically grounded expression control via AUs. Rather than relying on the network to infer geometry from AUs implicitly, we first predict a dynamic sequence of 2D facial landmarks conditioned on the input AUs and speech features. This sequence serves as a spatial-temporal scaffold that explicitly captures how muscle activations manifest as facial motion. We then employ a diffusion-based model to synthesize high-quality, temporally coherent video frames from these landmark trajectories, achieving expressive and controllable talking head generation that faithfully reflects the intended emotions. Fig.~\ref{fig:example} illustrates how our method can smoothly interpolate AUs across frames (indicated by the arrows).

Our experiments demonstrate that the proposed AU-to-landmark-to-video pipeline significantly enhances the realism, accuracy, and fine-grained controllability of generated facial expressions. Compared to FG-EmoTalk~\cite{FGEmoTalk}, our method achieves more precise facial geometry, smoother temporal transitions, and closer alignment with the intended AU inputs — highlighting the importance of explicit structural modeling for high-fidelity expression synthesis.

Our main contributions are:
\begin{itemize}
    %\item We propose an explicit AU-to-landmark prediction module to generate temporally consistent facial motion from input AUs.
    \item We introduce an explicit AU-to-landmark prediction module that generates temporally consistent and expressive facial motion, enabling fine-grained control over individual muscle activations.
    \item We develop a diffusion-based video generator conditioned on the predicted landmarks to synthesize realistic and controllable talking head videos.
    \item We show that our approach outperforms state-of-the-art methods in fine-grained expression accuracy, perceptual quality, and temporal coherence.
\end{itemize}

%Contribution
%1. First to tackle 18 action units for facial expression editing
%2. Model that predicts keypoint automatically
%3. Simple - only need action units, audios, and reference image  do not need a driving video (extracted keypoints)   Do not need clip text editing
\section{Related Work}
\label{sec:relatedwork}

% TODO: Video Diffusion Models

\subsection{Audio-driven Talking Head Generation}
\label{sec:relatedwork_audio}
Audio-driven talking head generation tasks try to synchronize lip movements with the driving speech audio while generating natural head posture and expression dynamics. Early methods integrate 3D face priors into GAN frameworks~\cite{wang2021audio2head, 10204743, 10.1145/3394171.3413532, 10.1007/978-3-031-19775-8_39, zhou2021pose}. Audio2Head~\cite{wang2021audio2head} and SadTalker~\cite{10204743} use 3DMM coefficients to guide adversarial generation for more accurate lip and pose synthesis. Wav2Lip~\cite{10.1145/3394171.3413532} introduces a dedicated lip-sync discriminator to dramatically improve mouth alignment in arbitrary videos. DFRF~\cite{10.1007/978-3-031-19775-8_39} leverages NeRF’s volumetric rendering for few-shot, audio-driven portrait synthesis. PC-AVS~\cite{zhou2021pose} models pose codes conditioned on audio and visual inputs to enable adjustable head rotation.

More recent research has turned to diffusion-based models, which provide more stable likelihood-based training and superior coverage of subtle lip and expression variations~\cite{10484496, ma2023dreamtalk, shen2023difftalk, 10.1007/978-3-031-73010-8_15, wang2025emotivetalk, xu2024hallo, wei2024aniportrait, zheng2024memo}. Difftalk~\cite{shen2023difftalk} improves robustness across diverse speakers by introducing a weakly supervised diffusion framework, while EMO~\cite{10.1007/978-3-031-73010-8_15} repurpose large-scale text-to-image diffusion backbones to achieve high-resolution face video generation. EMMN~\cite{10378627} integrates memory-guided temporal modules into diffusion models to capture long-term audio context for richer expressiveness and preserve identity consistency with smooth motion over extended sequences. In this work, we further leverage 2D facial keypoints and diffusion models to generate more realistic and high-fidelity talking head videos.

%----------------------------------------------------------------------
\subsection{Controllable Talking Head Generation}
\label{sec:controllable}
Controllable talking head generation remains a very challenging task due to the complex nature of facial motion and the need for precise, user‐driven control over subtle expression changes. Early attempts tackled this by conditioning on one-hot labels~\cite{10.1145/3528233.3530745, Gan_2023_ICCV, 10658061, 10.1109/TMM.2021.3099900, 10.1145/3072959.3073658, 10049691, 10458318, Sinha2022EmotionControllableGT, inproceedings}. EAMM~\cite{10.1145/3528233.3530745} produces one-shot emotional faces by transferring motion from an emotional driving video while masking the mouth to preserve lip sync. EAT~\cite{Gan_2023_ICCV} adapts a pre-trained audio-driven talking head model to produce discrete emotion styles. FlowVQTalker~\cite{10658061} models emotions as components in a normalizing flow over 3DMM expression coefficients and uses vector-quantized image generation to produce vivid, diverse emotional videos. EDTalk~\cite{tan2025edtalk} disentangles expression latents into neutral and emotion‐basis components, mixing learned emotion vectors with audio‐derived features to synthesize controllable emotional talking‐head videos. While these methods can produce distinct emotional categories, relying on a small set of coarse labels cannot capture the full nuance and blend of real human expressions.

Recent approaches tries to enable more detailed expression controls. While some methods transfer expressions from reference videos~\cite{10.1145/3664647.3681198, 10.1609/aaai.v37i2.25280, ma2023dreamtalk, wang2022pdfgc}, others leverage text prompts to guide their emotional video generation~\cite{tan2024style2talker, cui2025hallo, wang2025emotivetalk}. MEMO~\cite{zheng2024memo} detects emotion directly from audio and then uses a memory-guided temporal module to store and recall those emotion cues over long sequences, yielding more coherent and expressive results. More recently, researchers have moved toward Facial Action Coding System (FACS) for fine-grained expression control. ETAU~\cite{10687525} utilized facial Action Unit quantization to capture richer, fine‐grained expressions while FG-EmoTalk~\cite{FGEmoTalk} enables direct manipulation of individual muscle group intensities via AU inputs, reflecting real facial dynamics more faithfully. Unlike ETAU and FG-EmoTalk, which inject AU embeddings directly into their generators, our two-stage pipeline translates per-frame AU intensities into precise 2D facial landmarks. By grounding muscle activations in explicit landmarks, we are able to achieve more accurate and realistic facial movements.

\section{Method}
\label{sec:method}
\begin{figure*}[t]
  \centering
   \includegraphics[width=0.98\linewidth]{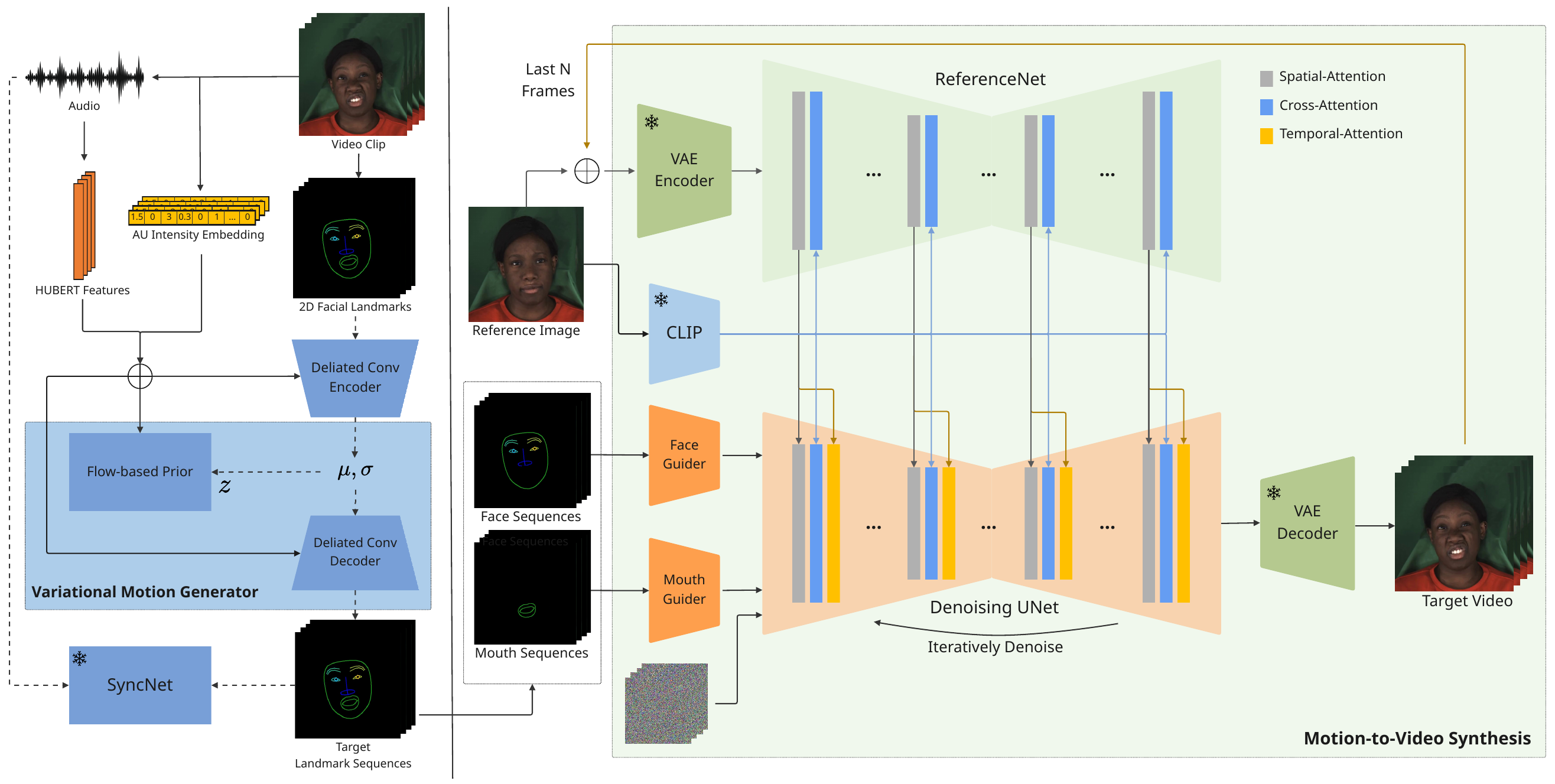}
   \caption{\textbf{Overview of our model.} Our method has two stages. In the first stage, our \textbf{Variational Motion Generator} extract audio features and the AU intensity embeddings, concatenating them into a dilated convolutional VAE encoder that produces per‐frame posterior parameters $\mu$ and $\sigma$. We then sample latent codes $z$ via a flow-based prior and decode them with a dilated convolutional decoder into the target landmark sequences. The sequences are then used in the second stage, \textbf{Motion-to-Video Synthesis}, where we seperate the landmarks into facial and mouth sequences. We utilize a diffusion-based model to transform the landmark sequences into the target video. Note: Dashed arrows means the process is only performed during training.}
   \label{fig:model_structure}
\end{figure*}

Fig.~\ref{fig:model_structure} the pipeline of our method, which is composed of two stages: (1) a Variational Motion Generator (VMG) that fuses audio features and AU intensity embeddings in a dilated-Conv VAE with a flow-based prior to produce expressive 2D landmark sequences (see Sec.~\ref{sec:method_3.2}); and (2) a diffusion-based Motion-to-Video Synthesis module that utilizes the landmark sequences predicted from the first stage to generate high-quality portrait videos with temporal stability (see Sec.~\ref{sec:method_3.3}).

\subsection{Preliminaries}
\label{sec:method_3.1}

\begin{figure}[t]
  \centering
   \includegraphics[width=0.95\linewidth]{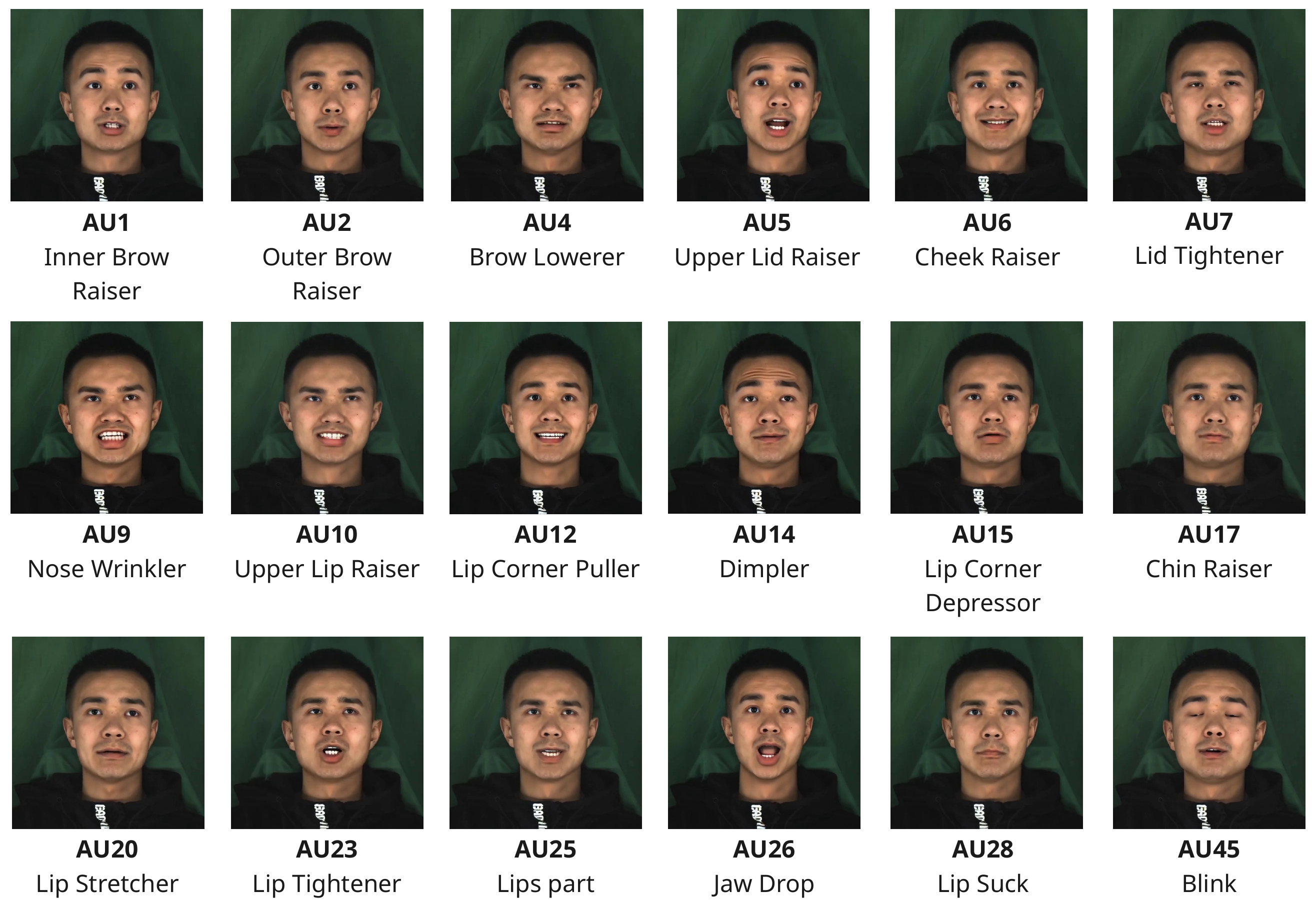}
   \caption{\textbf{Examples of our generated frames with each single AU.} Each AU typically conveys a specific facial movement which are accurately reflected in our generated outputs.  }
   \label{fig:au}
\end{figure}
% TODO: Is this figure good?

\subsubsection{Facial Action Unit}
\label{sec:method_3.1.1}
Facial Action Coding System (FACS), introduced by Ekman and Friesen~\cite{Ekman_1978_10190}, categorizes human facial movements by their visible changes on the face. The system encodes each underlying muscle contraction as an Action Unit (AU) by detecting subtle, instantaneous variations in facial appearance. FACS has become the standard for detailed expression analysis and synthesis because it covers both subtle and large‐scale facial movements. Moreover, AUs are frequently used to represent emotions. For example, “happiness” usually engages AU6 (Cheek Raiser) and AU12 (Lip Corner Puller), whereas “sadness” often involves AU4 (Brow Lowerer), AU7 (Lid Tightener), and AU15 (Lip Corner Depressor). Our model leverages 18 AUs detectable from OpenFace~\cite{amos2016openface}, as shown in Fig.~\ref{fig:au}.

\subsubsection{Latent Diffusion Models}
\label{sec:method_3.1.2}
Latent diffusion models operate in the latent space of an autoencoder rather than pixel space, enabling efficient and high-resolution generation. Concretely, given an image $I \in \mathbb{R}^{H_I \times W_I \times 3}$ and the conditioning embeddings $c$, we first encode $z_0 = \mathcal{E}(I)$ using an encoder $\mathcal{E}$ and decode $z_0$ to reconstruct the image via $x_{recon}=\mathcal{D}(z_0)$. During training, a time-conditioned UNet $\epsilon_\theta$ then learns to denoise latents $z_t$ back towards $z_0$ over $T$ steps, where
\begin{equation}
    z_t = \sqrt{\alpha_t}\,z_{t-1}
        + \sqrt{1-\alpha_t}\,\epsilon,
    \quad
    \epsilon \sim \mathcal{N}(0,1)
\end{equation}
The training objective is as follow:
\begin{equation}\label{eq:ldm-loss}
    \mathcal L
    = \mathbb E_{t,\epsilon}
    \bigl[||\epsilon - \epsilon_\theta(z_t,,c,,t)||_2^2\bigr]
\end{equation}
where at each step $t$, the network predicts the injected noise $\epsilon$ given the noisy latent, the timestep embedding, and the conditioning features $c$.

At inference time, we initialize $z_T \sim \mathcal{N}(0,I)$ and iteratively apply deterministic sampling methods (\emph{e.g.}\ DDPM~\cite{10.5555/3495724.3496298}, DDIM~\cite{song2021denoising}) to denoise $z_t$ to $z_0$ before decoding $z_0$ via the decoder $\mathcal D$.

\subsection{Variational Motion Generator}
\label{sec:method_3.2}
We introduce a conditional variational autoencoder (VAE) that learns to map speech audio and facial Action Unit intensities into expressive 2D landmark trajectories, allowing diverse and natural head motion synthesis.

\subsubsection{Input and Output Representations}
\label{sec:method_3.2.1}
To better capture semantic and fine-grained expression cues, we extract frame-level audio embeddings $a_t$ from the raw waveform using a pretrained HuBERT ASR model~\cite{10.1109/TASLP.2021.3122291}. Facial expression information is represented by an 18-dimensional AU intensity vector obtained from OpenFace~\cite{7284869}, where each entry encodes the type and intensity (ranging from 0 to 5) of a specific facial action unit. For motion supervision, we extract 162 crucial facial keypoints (covering eyebrows, eyes, lips, nose, and face contour) from the reconstructed 3D facial landmarks~\cite{48292}. These keypoints serve as the target landmark trajectories for our variational motion generator.
% TODO: Add intensity figure

\subsubsection{Audio-to-Motion VAE}
\label{sec:method_3.2.2}
Inspired by GeneFace~\cite{ye2023geneface}, we implement our motion generator as a single conditional VAE that both encodes and decodes the entire landmark sequences with long‐range temporal context. We process the concatenated HuBERT audio embeddings $a_t$ and the AU intensity embeddings $u_t$ with a stack of 1D dilated convolutions, producing per-frame mean $\mu_t$ and log-variance vectors $\sigma_t$ for the input latent code $z_t$:
\begin{equation}
    q_{\phi}\bigl(z_{t}\mid L_{t},c_{t}\bigr)
    = \mathcal{N}\bigl(z_{t};\,\mu_{t},\,\mathrm{diag}\bigl(\exp(\sigma_{t})^{2}\bigr)\bigr)\ ,
\end{equation}
where $L_{t}$ is the ground truth landmarks and $c_t = [a_t; u_t]$ is the audio-AU conditioning at each frame $t$, and $\phi$ denotes the encoder parameters. 
To capture the strong temporal correlations inherent in facial motion and avoid overly smooth posteriors, we also use a normalizing flow-based prior $p_{\epsilon}(z_t \mid c_t)$ composed of four residual coupling layers to compute the ELBO’s KL term against this learned flow distribution (See Supplementary for more detail).
% the convolutional layers have incrementally increased dilation factors that allow its receptive field to grow exponentially with depth.

\subsubsection{Loss Function}
\label{sec:method_3.2.3}
With $L$ and $\hat L$ being the ground truth and the predicted 2D landmarks respectively, we train our audio-to-motion VAE model with an MSE reconstruction loss
\begin{equation}
    \mathcal{L}_{\mathrm{MSE}}
    = \mathbb{E}_{z\sim q_\phi(z\mid L,c)}\bigl[\| L - \hat L_\theta\|_2^2\bigr]
\end{equation}
and a continuity loss
\begin{equation}
    \mathcal{L}_{\mathrm{cont}}
    = \mathbb{E}_t\bigl\|\hat L_{t,\theta} - \hat L_{t-1,\theta}\|_2^2
\end{equation}
to encourage accurate reconstruction and temporal smoothness. Since the flow-based prior breaks the closed-form ELBO, we train the VAE model with the Monte Carlo ELBO following~\cite{10.5555/3540261.3541331}:
\begin{equation}
    \mathcal{L}_{\mathrm{KL}}
    = \mathrm{KL}\bigl(q_\phi(z\mid L,c)\,\|\,p_\epsilon(z\mid c)\bigr),
\end{equation}
where $\phi,\theta,\epsilon$ denote the encoder, decoder, and prior parameters respectively, and $c = [a; u]$ is the audio–AU conditioning.

% syncNet detail in supplementary?
We also independently train a \textit{sync-expert} $D_{\mathrm{sync}}$ that measures the possibility of the synchronization between input audio and 2D landmarks:
\begin{equation}
    \mathcal{L}_{\mathrm{sync}}
    = 1 - \cos\bigl(D_{\mathrm{sync}}(\hat L),\,D_{\mathrm{sync}}(a)\bigr).
\end{equation}
Unlike GeneFace, which uses a simple BCE loss for $\mathcal{L}_{\mathrm{sync}}$, we replace it with cosine embedding loss to keep the gradients well-conditioned (see Supplementary for more details). Our loss is then as follows:

\begin{align}
    \mathcal{L}_{\mathrm{VAE}}
    &= \lambda_{\mathrm{MSE}}\,\mathcal{L}_{\mathrm{MSE}}
      + \lambda_{\mathrm{KL}}\,\mathcal{L}_{\mathrm{KL}}
      \notag\\
    &\quad
      + \lambda_{\mathrm{cont}}\,\mathcal{L}_{\mathrm{cont}}
      + \lambda_{\mathrm{sync}}\,\mathcal{L}_{\mathrm{sync}}\,.
\end{align}

\begin{table*}
  \centering
  \begin{tabular}{@{}lcccccc@{}}
    \toprule
    Method & PSNR($\uparrow$) & SSIM($\uparrow$) & FID($\downarrow$) & Sync($\uparrow$) & M-LMD/F-LMD($\downarrow$) & $Acc_{emo}$($\uparrow$) \\
    \midrule
    EAMM~\cite{10.1145/3528233.3530745} & 18.81 & 0.5289 & 81.88 & 1.125 & 3.722/7.343 & 31.74$\%$ \\
    EAT~\cite{Gan_2023_ICCV} & 21.75 & 0.6780 & 20.21 & \textbf{2.054} & 2.756/3.099 & 70.40$\%$ \\
    MEMO~\cite{zheng2024memo} & \textbf{22.05} & 0.6530 & 33.31 & 1.008 & 2.935/3.978 & 48.6$\%$ \\
    \midrule
    \textbf{Ours} & 22.01 & \textbf{0.6830} & \textbf{18.00} & 1.688 & \textbf{2.288}/\textbf{2.857} & \textbf{78.03$\%$} \\
    %\midrule
    %Ground Truth & $\infty$ & 1.0 & 0 & 2.226 & 0.0/0.0 & 86$\%$ \\
    \bottomrule
  \end{tabular}
  \caption{\textbf{Quantitative comparisons with state-of-the-art methods on MEAD.} The symbols ‘$\uparrow$’ and ‘$\downarrow$’ indicate higher and lower metric values for better results, respectively.}
  \label{tab:quantitative}
\end{table*}
% TODO: Should we add ground truth?

\subsection{Motion-to-Video Synthesis}
\label{sec:method_3.3}
In this section, we introduce a Motion-to-Vidoe Synthesis framework that turns the landmark sequences produced by VMG from the first stage into full‐resolution, lip‑synced talking‑head videos that injects appearance, pose and temporal cues into a diffusion‐based backbone.

\subsubsection{Network Architecture}
\label{sec:method_3.3.1}
\textbf{Overview.} Our model is built on a latent diffusion backbone identical to Stable Diffusion 1.5, a pretrained VQ-VAE encoder/decoder with a UNet denoiser, but replace the text conditioning with audio and motion signals. Additionally, our method consists of three key components: 1) ReferenceNet, which extracts identity and appearance features from a reference image; 2) Pose Guider, which encodes audio-synced landmarks to guide pose and lip motion; 3) Temporal Module, which enforces smooth frame-to-frame continuity through lightweight temporal attention.

\noindent \textbf{ReferenceNet.} ReferenceNet is designed to inject high-fidelity appearance details from the reference image into the diffusion backbone. It is a UNet architecture with the same number of layers and block structure as the denoising model from Stable Diffusion. During training, the first frame is passed through ReferenceNet, which produces multi-scale feature maps. These maps are fused into the corresponding UNet layers via spatial-attention: each feature map from ReferenceNet is t-fold concatenated to match the temporal dimension of the noisy latents, then self-attention is applied and the first half of the channels retained. This alignment ensures that the model selectively incorporates fine-grained textures and background details.

\noindent \textbf{Pose Guider.} We introduce two lightweight Pose Guiders - one encoding only the mouth keypoints and another encoding the remaining facial keypoints. Both guiders reshape keypoint maps to match UNet’s latent resolution and process the maps through four 4×4-kernel, stride-2 convolutional layers with 16, 32, 64, and 128 channels (following ~\cite{hu2024animate}). The outputs are then added directly to the noisy latent before entering the denoiser, enabling focused lip-sync refinement from the mouth guider while the other guider governs broader expression and head pose. The final projection layer is zero convolution so that they don’t perturb the pretrained denoiser at training start.

\noindent \textbf{Temporal Alignment.} In diffusion‐based video generation, it is crucial to preserve coherence and consistency across frames. Following~\cite{xu2024hallo}, we select a small set of previously generated frames and concatenate their noisy latents along the temporal axis. Within each Res-Trans block, immediately after spatial and cross-attention components, we reshape the feature map $x \in \mathbb{R}^{b \times t \times h \times w \times c}$ to $x \in \mathbb{R}^{(b \cdot h \cdot w) \times t \times c}$ and apply self-attention over the $t$-dimension. The attended output is then reshaped back and added via a residual connection. This temporal attention enforces smooth frame-to-frame consistency in appearance and motion, eliminating flicker and reducing the need for separate, complex temporal priors.

\begin{figure*}[t]
  \centering
   \includegraphics[width=0.99\linewidth]{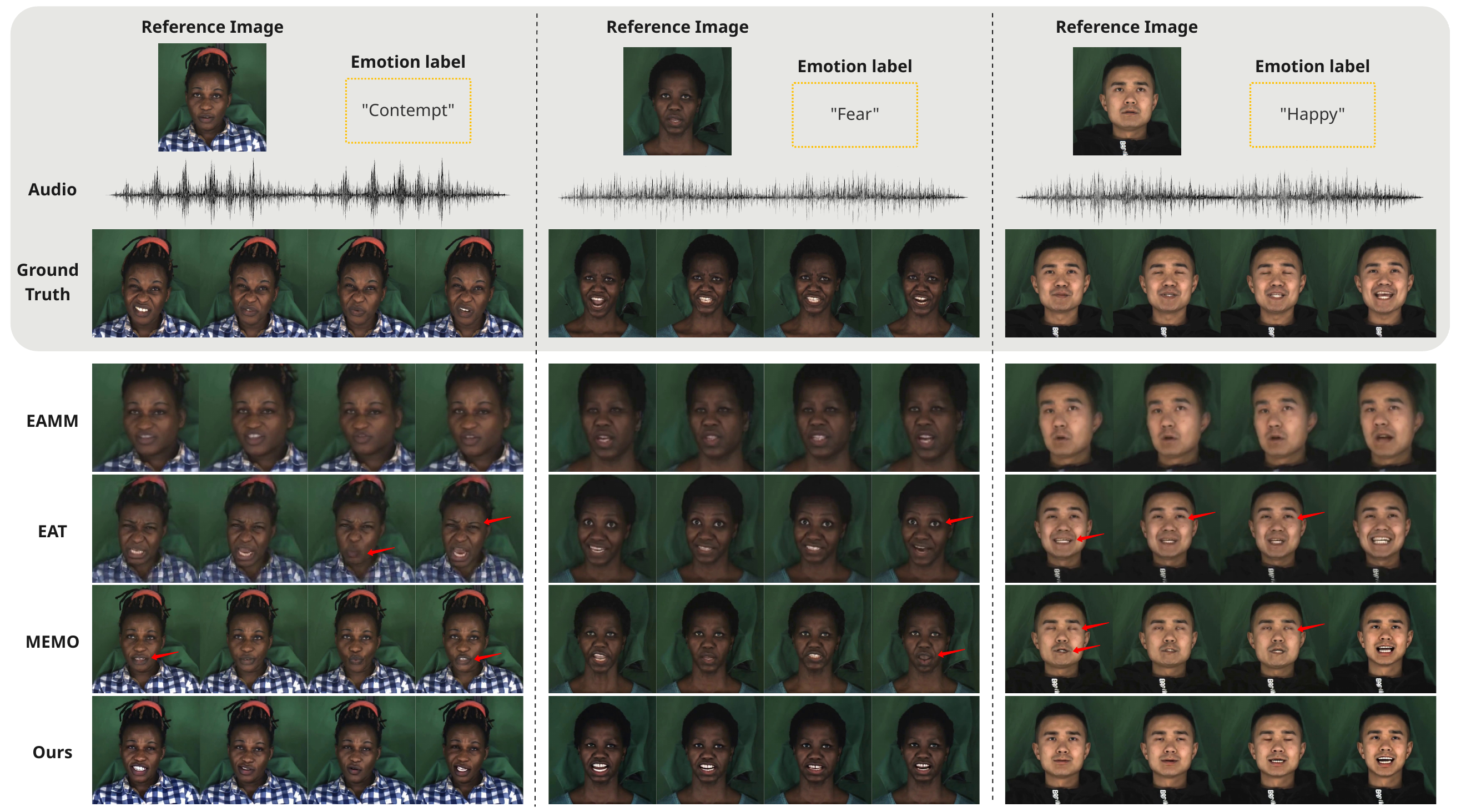}
   \caption{\textbf{Qualitative comparisons with state-of-the-art methods.} Our method generates temporally consistent and visually realistic videos with fine-grained and accurate facial expressions. In contrast, competing methods often produce noticeable artifacts, inconsistencies, or fail to reflect the intended emotional cues accurately (see red arrows for examples). This highlights the advantage of our explicit AU-based control mechanism. }
   \label{fig:qualitative}
\end{figure*}

\subsubsection{Training and Inference}
\label{sec:method_3.3.2}
The training process consists of two phases. First, we train on individual frames to establish high‐quality single‐image synthesis conditioned on appearance and pose. We temporarily exclude the temporal layer and randomly sample one frame from each video clip as the reference image in this stage. The VAE encoder and decoder remain frozen, while UNet and ReferenceNet are trained and initialized with Stable Diffusion 1.5 weights. The Pose Guider is initialized from ControlNetMediaPipeFace\footnote{\url{https://huggingface.co/CrucibleAI/ControlNetMediaPipeFace}}, which is a ControlNet structure trained on 3D facial keypoints~\cite{48292}. In the second phase, we introduce the temporal layer, initialized from AnimateDiff’s~\cite{guo2023animatediff} pretrained motion module, into the Res-Trans blocks. The input for the model are 16-frame video clips. We freeze all other weights to just focus on the temporal consistency. %Here, the model learns to leverage the new self-attention over the temporal axis so that generated sequences exhibit smooth, coherent motion across extended video clips.
\section{Experiments}
\label{sec:exp}
\subsection{Experimental Setup}
\label{sec:exp_4.1}
\subsubsection{Datasets}
\label{sec:exp_4.1.1}
We use the MEAD dataset to train both Variational Motion Generator and diffusion‐based Motion-to-Video Synthesis stages. MEAD contains high‐quality emotional talking‐head videos of 60 professional actors speaking under 8 emotion categories (anger, disgust, fear, happiness, sadness, surprise, contempt, and neutral). For SOTA comparison, we select $1,000$ MEAD videos for evaluation, with equal number from each emotion category. In addition, we take the first frame from the same actor’s neutral video as the source image to ensure consistent identity reference.

\subsubsection{Implementation Details}
\label{sec:exp_4.1.2}
All videos are processed at 25 FPS with audio sampled at 16 kHz, and frames are center-cropped and resized to 512 × 512. Both modules are run on 2 NVIDIA RTX A6000 GPUs. For the Variational Motion Generator, we train for 40k steps with a batch size of 2,000 clips and 16 sampled frames per clip. We use Adam Optimization with a learning rate of 1e-4. The hyperparameters of the ELBO terms $\lambda_{\mathrm{MSE}}$, $\lambda_{\mathrm{KL}}$, $\lambda_{\mathrm{cont}}$, and $\lambda_{\mathrm{sync}}$ were set to 5.0, 0.5, 3.0, and 0.01 respectively.

As for the Motion-to-Video module, we train the first phase using a batch size of 8 and batch size of 1 with 16 frames per clip in the second phase. To reduce artifacts and improve accuracy, all keypoint sequences generated from VMG are first rigidly aligned to the reference image before synthesis. Additionally, we concatenate the motion module’s latents with the last 2 ground-truth frames to ensure continuity in the second stage. Both stages use a fixed learning rate of 1e-5 for 30k steps and randomly drop the reference image and motion frames with probability $0.1$ during training. At inference time, we run 40 DDIM steps to sample the final video sequences.

\subsubsection{Baselines}
\label{sec:exp_4.1.3}
We compare with several state-of-the-art talking head generation models that use discrete emotion labels: 1) EAMM~\cite{10.1145/3528233.3530745} synthesizes one‑shot emotional talking heads by transferring motion from an emotion‑driving video; 2) EAT~\cite{Gan_2023_ICCV} adapts a pretrained audio‑driven talking‑head model by injecting learnable emotion adapters; 3) MEMO~\cite{zheng2024memo} is an audio‑driven diffusion framework that first detects emotion from the input audio and then refines facial expressions via an emotion‑aware layer norm. For fair comparison, we reconstruct videos using AU embeddings extracted from the ground truth. By driving our model with the same AU patterns that correspond to each emotion label, we align our input conditions with those of the label-based methods.
%Emotion: VideoReTalking,
%video transfer: StyleTalk, AniPortrait, V-Express
%text prompt: EAT, Style2Talker, EmotiveTalk, Hallo2

\begin{figure*}[t]
  \centering
   \includegraphics[width=0.95\linewidth]{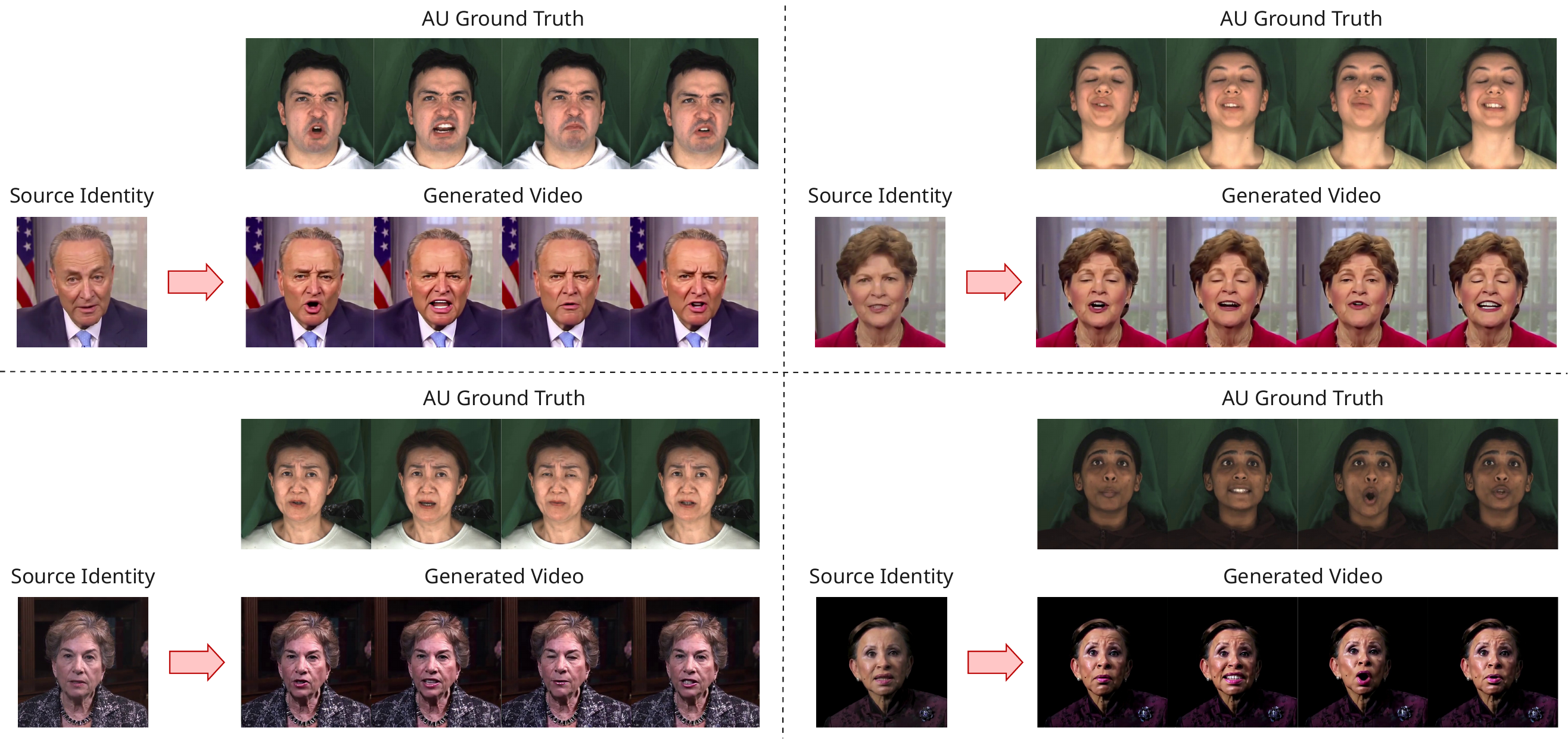}
   \caption{Additional qualitative results using different source identities from the HDTF dataset.}
   \label{fig:hdtf_example}
\end{figure*}

\subsubsection{Evaluation Metrics}
\label{sec:exp_4.1.4}
We access the results using metrics that evaluate three aspects: image‐quality, audio‐visual synchronization, and emotion‐accuracy. For image generation quality, we utilize SSIM~\cite{1284395}, PSNR, and FID~\cite{10.5555/3295222.3295408}. We evaluate audio-visual synchronization using M-LMD for lip movement accuracy and F-LMD~\cite{chen2019hierarchical} for full face alignment. In addition, we use SyncNet~\cite{Chung16a} confidence score to measure the synchronization of lip motion with input audio. To ensure accurate evaluations, we crop and align the faces before calculating above metrics, following EAT~\cite{Gan_2023_ICCV}. Lastly, emotional accuracy $Acc_{emo}$ is computed by running a fine‐tuned Emotion‐Fan~\cite{meng2019frame} classifier on the generated clips, using the MEAD training set for adaptation. To evaluate $Acc_{emo}$ for our AU‑driven conditioning, we map each emotion label to its canonical AU combination:
\begin{itemize}
    \item Angry: AU4 (Brow Lower) + AU7 (Lid Tightener) + AU10 (Upper Lip Raiser) + AU23 (Lip Tightener)
    \item Disgust: AU7 (Lid Tightener) + AU14 (Dimpler) + AU17 (Chin Raiser)
    \item Fear: AU2 (Outer Brow Raiser) + AU4 (Brow Lower) + AU5 (Upper Lid Raiser) + AU7 (Lid Tightener) + AU26 (Jaw Drop)
    \item Happy: AU6 (Cheek Raiser) + AU12 (Lip Corner Puller)
    \item Sad: AU4 (Brow Lower) + AU7 (Lid Tightener) + AU15 (Lip Corner Depressor)
    \item Surprised: AU1 (Inner Brow Raiser) + AU2 (Outer Brow Raiser) + AU5 (Upper Lid Raiser) + AU26 (Jaw Drop)
    \item Contempt: AU12 (Lip Corner Puller) + AU14 (Dimpler)
\end{itemize}
This ensures a fair comparison that aligns our AU‑based driving scheme with other emotional talking head generation methods and more faithfully reflects fine‑grained expressive accuracy.

\subsection{Quantitative Results}
Tab.~\ref{tab:quantitative} shows the quantitative comparison of our method with SOTA methods on MEAD dataset. Overall, our method outperforms other methods in most of the evaluation metrics, indicating our superior capabilities of generating high-fidelity, faithful, and vivid emotional expressions. Notably, our especially high $Acc_{emo}$ score demonstrates that leveraging explicit AUs enables richer and more accurate facial expression dynamics.
% TODO: Mention why eat has higher sync score.
% TODO: Mention we have lower sync score than EAT is because we use landmarks as driving force at second stage, which reduce a little bit of the sync effect (limitation)?

\begin{table*}
  \centering
  \begin{tabular}{@{}lcccccc@{}}
    \toprule
    Method & PSNR($\uparrow$) & SSIM($\uparrow$) & FID($\downarrow$) & Sync($\uparrow$) & M-LMD/F-LMD($\downarrow$) & $Acc_{emo}$($\uparrow$) \\
    \midrule
    One-hot Encode & 21.60 & 0.6673 & 29.38 & 1.481 & 2.568/3.165 & 47.6$\%$ \\
    Audio-driven & 21.07 & 0.6330 & 32.47 & 1.476 & 2.761/3.540 & 48.2$\%$ \\
    \midrule
    \textbf{Ours (AU)} & \textbf{22.01} & \textbf{0.6830} & \textbf{18.00} & \textbf{1.688} & \textbf{2.288}/\textbf{2.857} & \textbf{78.03$\%$} \\
    \bottomrule
  \end{tabular}
  \caption{\textbf{Ablation study.} We compare between using AU vector and one-hot emotion labels or audio-driven embedding. It can be seen that AUs play a key role to convey accurate emotional expression in the generated outputs.}
  % TODO: Say that our methods are better than other two ablation study methods.
  \label{tab:ablation}
\end{table*}

\subsection{Qualitative Results}
\label{sec:exp_4.2}
\subsubsection{Qualitative Comparison}
\label{sec:exp_4.2.1}
Fig.~\ref{fig:qualitative} shows the visual comparisons of our method against SOTA models on MEAD examples. EAMM struggles to produce high-quality and expressive videos. Their results are often blurry and lack the sharp muscle contractions that define each expression. EAT delivers much clearer, more expressive results, but its reliance on just eight fixed emotion labels makes every emotion enactment virtually identical. For example, in the first case in Fig.~\ref{fig:qualitative}, EAT always lowers the eyebrows for ``contempt'' even though the ground truth shows raised brows. In the second case in Fig.~\ref{fig:qualitative}, it amplifies eye-widening for ``fear'' which does not match the source expression. MEMO generates results with higher quality than EAMM and EAT due to its diffusion-based architechture, but its audio-to-emotion structure is too coarse to capture subtle, AU-level variations as shown in all three examples in Fig.~\ref{fig:qualitative}. On the contrary, our AU-driven approach reconstruct the full range of muscle dynamics with high fidelity. We are able to recover exact lip shapes — whether it's parting lips (AU25), closing lips (AU20), or bared-teeth expression (AU10). In addition, we utilize AU45 to match the timing and duration of each blink with the ground truth as shown in the third case in Fig.~\ref{fig:qualitative}, while EAMM and EAT rarely blink and MEMO blinks randomly.

\subsubsection{Additional Qualitative Results}
\label{sec:exp_4.2.2}
To further validate our model’s generalization ability, we test it on HDTF dataset~\cite{zhang2021flow}, a large collection of talking‑head videos collected from YouTube. Since HDTF does not have emotion annotations, we extract per‑frame AU intensity sequences from MEAD dataset as ground‑truth clips. We then use those AU embeddings to drive source identities through our two‑stage pipeline. As shown in Figure~\ref{fig:hdtf_example}, our method successfully transfers a wide range of fine‑grained facial expressions. For example, the brow lowering (AU4), lip corner depressing (AU15), and nose wrinkling (AU9) in the upper-left example; lip corner pulling (AU12), chin raising (AU17), and blinking (AU45) in the upper-right example; lip sucking (AU28) and eyelid tightening (AU7) in the bottom-left example; upper eyelid tightening (AU7), lip stretching (AU20), and lip tightening (AU23) in the bottom-right example.

% TODO: Mention single AU.
% TODO: intensity

\subsection{Ablation Study}
% Does this ablation study make sense?
% TODO: qualitative results?
To assess the effectiveness and importance of our fine‐grained AU conditioning, we conduct two ablation studies in which we replace our AU intensity embeddings in the Variational Motion Generation stage with coarser emotion representations.

\noindent \textbf{One-Hot Emotion Labels.} We swap out our AU intensity vector for an 8-dimensional one-hot encoding matching the discrete emotion labels used by MEAD dataset.

\noindent \textbf{Audio-Driven Emotion Embeddings.} In this study, we mimic MEMO's in-line emotion conditioning - we train a lightweight classifier atop our frame-wise audio features. Specifically, we add a two-layer MLP with a softmax that map each audio vector to one of the eight emotion categories. At inference, the AU vector is then replaced by the predicted emotion probability embeddings as the VAE conditioning input.
%We train with an additional cross-entropy against MEAD’s emotion labels.

\noindent \textbf{Results.} Tab.~\ref{tab:ablation} shows that replacing our fine‐grained AU vectors with either one‐hot emotion labels or audio‐driven emotion embeddings leads to a severe deterioration in anatomical detail. In particular, both ablations suffer a substantial drop in $Acc_{emo}$, confirming that without per‐unit intensity signals the model cannot reproduce the nuanced Action Unit patterns that our metric is designed to evaluate.
%Only our AU-driven conditioning delivers both high visual fidelity and faithful, fine-grained expression dynamics.
\section{Conclusion}
In this paper, we present a novel two-stage framework for expressive, fine‑grained audio-driven talking‑head generation conditioned on per‑frame Action Unit embeddings. In the first stage, our Variational Motion Generator fuses audio features and AU intensity embeddings in a dilated convolutional VAE with a flow‑based prior to predict temporally coherent 2D landmark sequences. In the second stage, we leverage the predicted landmark sequences generated from the first stage to synthesize videos with a diffusion-based architecture. Experiments demonstrate that our AU‑driven pipeline outperforms state‑of‑the‑art methods across image quality, audio‑visual synchronization, and fine-grained expression control.

\noindent \textbf{Limitations and Future Work.} While our method achieves superior performance in both quantitative and qualitative evaluations, several limitations remain. First, our approach relies on accurate AU extraction from tools like OpenFace, which can be sensitive to head pose or occlusions. Incorporating 3D landmark estimation or self-supervised AU detection may address this.
Second, our method does not currently support user-level personalization of expression style — the same AU intensities yield similar motions across identities. Future work could explore style tokens or disentangled personality embeddings to model individual expression tendencies. Lastly, while MEAD offers a controlled dataset, real-world deployment requires handling in-the-wild variability such as emotion ambiguity, occlusions, and non-verbal vocalizations like laughter. Expanding training with in-the-wild datasets may further enhance robustness.
% TODO: future work. In future work, we plan to integrate richer anatomical priors and interactive control modalities, explore lightweight architectures for real‑time deployment, and extend our approach to full‑body or multi‑person scenarios.
{
    \small
    \bibliographystyle{ieeenat_fullname}
    \bibliography{main}

\begin{thebibliography}{48}
\providecommand{\natexlab}[1]{#1}
\providecommand{\url}[1]{\texttt{#1}}
\expandafter\ifx\csname urlstyle\endcsname\relax
  \providecommand{\doi}[1]{doi: #1}\else
  \providecommand{\doi}{doi: \begingroup \urlstyle{rm}\Url}\fi

\bibitem[Amos et~al.(2016)Amos, Ludwiczuk, and Satyanarayanan]{amos2016openface}
Brandon Amos, Bartosz Ludwiczuk, and Mahadev Satyanarayanan.
\newblock Openface: A general-purpose face recognition library with mobile applications.
\newblock Technical report, CMU-CS-16-118, CMU School of Computer Science, 2016.

\bibitem[Baltrušaitis et~al.(2015)Baltrušaitis, Mahmoud, and Robinson]{7284869}
Tadas Baltrušaitis, Marwa Mahmoud, and Peter Robinson.
\newblock Cross-dataset learning and person-specific normalisation for automatic action unit detection.
\newblock In \emph{2015 11th IEEE International Conference and Workshops on Automatic Face and Gesture Recognition (FG)}, pages 1--6, 2015.

\bibitem[Chen et~al.(2019)Chen, Maddox, Duan, and Xu]{chen2019hierarchical}
Lele Chen, Ross~K Maddox, Zhiyao Duan, and Chenliang Xu.
\newblock Hierarchical cross-modal talking face generation with dynamic pixel-wise loss.
\newblock In \emph{Proceedings of the IEEE Conference on Computer Vision and Pattern Recognition}, pages 7832--7841, 2019.

\bibitem[Chung and Zisserman(2016)]{Chung16a}
J.~S. Chung and A. Zisserman.
\newblock Out of time: automated lip sync in the wild.
\newblock In \emph{Workshop on Multi-view Lip-reading, ACCV}, 2016.

\bibitem[Cui et~al.(2025)Cui, Li, Yao, Zhu, Shang, Cheng, Zhou, Zhu, and Wang]{cui2025hallo}
Jiahao Cui, Hui Li, Yao Yao, Hao Zhu, Hanlin Shang, Kaihui Cheng, Hang Zhou, Siyu Zhu, and Jingdong Wang.
\newblock Hallo2: Long-duration and high-resolution audio-driven portrait image animation.
\newblock In \emph{The Thirteenth International Conference on Learning Representations}, 2025.

\bibitem[Ekman and Friesen(1978)]{Ekman_1978_10190}
P. Ekman and W.V. Friesen.
\newblock \emph{Facial action coding system: A technique for the measurement of facial movement}.
\newblock Consulting Psychologists Press, Palo Alto, CA, 1978.

\bibitem[Eskimez et~al.(2022)Eskimez, Zhang, and Duan]{10.1109/TMM.2021.3099900}
Sefik~Emre Eskimez, You Zhang, and Zhiyao Duan.
\newblock Speech driven talking face generation from a single image and an emotion condition.
\newblock \emph{Trans. Multi.}, 24:\penalty0 3480–3490, 2022.

\bibitem[Gan et~al.(2023)Gan, Yang, Yue, Sun, and Yang]{Gan_2023_ICCV}
Yuan Gan, Zongxin Yang, Xihang Yue, Lingyun Sun, and Yi Yang.
\newblock Efficient emotional adaptation for audio-driven talking-head generation.
\newblock In \emph{Proceedings of the IEEE/CVF International Conference on Computer Vision (ICCV)}, pages 22634--22645, 2023.

\bibitem[Guo et~al.(2024)Guo, Yang, Rao, Liang, Wang, Qiao, Agrawala, Lin, and Dai]{guo2023animatediff}
Yuwei Guo, Ceyuan Yang, Anyi Rao, Zhengyang Liang, Yaohui Wang, Yu Qiao, Maneesh Agrawala, Dahua Lin, and Bo Dai.
\newblock Animatediff: Animate your personalized text-to-image diffusion models without specific tuning.
\newblock \emph{International Conference on Learning Representations}, 2024.

\bibitem[Heusel et~al.(2017)Heusel, Ramsauer, Unterthiner, Nessler, and Hochreiter]{10.5555/3295222.3295408}
Martin Heusel, Hubert Ramsauer, Thomas Unterthiner, Bernhard Nessler, and Sepp Hochreiter.
\newblock Gans trained by a two time-scale update rule converge to a local nash equilibrium.
\newblock In \emph{Proceedings of the 31st International Conference on Neural Information Processing Systems}, page 6629–6640, Red Hook, NY, USA, 2017. Curran Associates Inc.

\bibitem[Ho et~al.(2020)Ho, Jain, and Abbeel]{10.5555/3495724.3496298}
Jonathan Ho, Ajay Jain, and Pieter Abbeel.
\newblock Denoising diffusion probabilistic models.
\newblock In \emph{Proceedings of the 34th International Conference on Neural Information Processing Systems}, Red Hook, NY, USA, 2020. Curran Associates Inc.

\bibitem[Hsu et~al.(2021)Hsu, Bolte, Tsai, Lakhotia, Salakhutdinov, and Mohamed]{10.1109/TASLP.2021.3122291}
Wei-Ning Hsu, Benjamin Bolte, Yao-Hung~Hubert Tsai, Kushal Lakhotia, Ruslan Salakhutdinov, and Abdelrahman Mohamed.
\newblock Hubert: Self-supervised speech representation learning by masked prediction of hidden units.
\newblock \emph{IEEE/ACM Trans. Audio, Speech and Lang. Proc.}, 29:\penalty0 3451–3460, 2021.

\bibitem[Hu(2024)]{hu2024animate}
Li Hu.
\newblock Animate anyone: Consistent and controllable image‐to‐video synthesis for character animation.
\newblock In \emph{Proceedings of the IEEE/CVF Conference on Computer Vision and Pattern Recognition (CVPR)}, 2024.

\bibitem[Ji et~al.(2022)Ji, Zhou, Wang, Wu, Wu, Xu, and Cao]{10.1145/3528233.3530745}
Xinya Ji, Hang Zhou, Kaisiyuan Wang, Qianyi Wu, Wayne Wu, Feng Xu, and Xun Cao.
\newblock Eamm: One-shot emotional talking face via audio-based emotion-aware motion model.
\newblock In \emph{ACM SIGGRAPH 2022 Conference Proceedings}, New York, NY, USA, 2022. Association for Computing Machinery.

\bibitem[Karras et~al.(2017)Karras, Aila, Laine, Herva, and Lehtinen]{10.1145/3072959.3073658}
Tero Karras, Timo Aila, Samuli Laine, Antti Herva, and Jaakko Lehtinen.
\newblock Audio-driven facial animation by joint end-to-end learning of pose and emotion.
\newblock \emph{ACM Trans. Graph.}, 36\penalty0 (4), 2017.

\bibitem[Liu et~al.(2024)Liu, Chen, Fan, Du, Chen, Chen, and Yu]{10.1145/3664647.3681198}
Tao Liu, Feilong Chen, Shuai Fan, Chenpeng Du, Qi Chen, Xie Chen, and Kai Yu.
\newblock Anitalker: Animate vivid and diverse talking faces through identity-decoupled facial motion encoding.
\newblock In \emph{Proceedings of the 32nd ACM International Conference on Multimedia}, page 6696–6705, New York, NY, USA, 2024. Association for Computing Machinery.

\bibitem[Lugaresi et~al.(2019)Lugaresi, Tang, Nash, McClanahan, Uboweja, Hays, Zhang, Chang, Yong, Lee, Chang, Hua, Georg, and Grundmann]{48292}
Camillo Lugaresi, Jiuqiang Tang, Hadon Nash, Chris McClanahan, Esha Uboweja, Michael Hays, Fan Zhang, Chuo-Ling Chang, Ming Yong, Juhyun Lee, Wan-Teh Chang, Wei Hua, Manfred Georg, and Matthias Grundmann.
\newblock Mediapipe: A framework for perceiving and processing reality.
\newblock In \emph{Third Workshop on Computer Vision for AR/VR at IEEE Computer Vision and Pattern Recognition (CVPR) 2019}, 2019.

\bibitem[Lyu et~al.(2024)Lyu, Lan, Hu, Jiang, Gan, and Xue]{10687525}
Jiayi Lyu, Xing Lan, Guohong Hu, Hanyu Jiang, Wei Gan, and Jian Xue.
\newblock Etau: Towards emotional talking head generation via facial action unit.
\newblock In \emph{2024 IEEE International Conference on Multimedia and Expo (ICME)}, pages 1--6, 2024.

\bibitem[Ma et~al.(2023{\natexlab{a}})Ma, Wang, Hu, Fan, Lv, Ding, Deng, and Yu]{10.1609/aaai.v37i2.25280}
Yifeng Ma, Suzhen Wang, Zhipeng Hu, Changjie Fan, Tangjie Lv, Yu Ding, Zhidong Deng, and Xin Yu.
\newblock Styletalk: one-shot talking head generation with controllable speaking styles.
\newblock In \emph{Proceedings of the Thirty-Seventh AAAI Conference on Artificial Intelligence and Thirty-Fifth Conference on Innovative Applications of Artificial Intelligence and Thirteenth Symposium on Educational Advances in Artificial Intelligence}. AAAI Press, 2023{\natexlab{a}}.

\bibitem[Ma et~al.(2023{\natexlab{b}})Ma, Zhang, Wang, Wang, Zhang, and Deng]{ma2023dreamtalk}
Yifeng Ma, Shiwei Zhang, Jiayu Wang, Xiang Wang, Yingya Zhang, and Zhidong Deng.
\newblock Dreamtalk: When expressive talking head generation meets diffusion probabilistic models.
\newblock \emph{arXiv preprint arXiv:2312.09767}, 2023{\natexlab{b}}.

\bibitem[Meng et~al.(2019)Meng, Peng, Wang, and Qiao]{meng2019frame}
Debin Meng, Xiaojiang Peng, Kai Wang, and Yu Qiao.
\newblock frame attention networks for facial expression recognition in videos.
\newblock In \emph{2019 IEEE International Conference on Image Processing (ICIP)}, pages 3866--3870. IEEE, 2019.

\bibitem[Pan et~al.(2023)Pan, Zhang, Cheng, Tan, Ding, Mitchell, and Yang]{10049691}
Ye Pan, Ruisi Zhang, Shengran Cheng, Shuai Tan, Yu Ding, Kenny Mitchell, and Xubo Yang.
\newblock Emotional voice puppetry.
\newblock \emph{IEEE Transactions on Visualization and Computer Graphics}, 29\penalty0 (5):\penalty0 2527--2535, 2023.

\bibitem[Pan et~al.(2024)Pan, Tan, Cheng, Lin, Zeng, and Mitchell]{10458318}
Ye Pan, Shuai Tan, Shengran Cheng, Qunfen Lin, Zijiao Zeng, and Kenny Mitchell.
\newblock Expressive talking avatars.
\newblock \emph{IEEE Transactions on Visualization and Computer Graphics}, 30\penalty0 (5):\penalty0 2538--2548, 2024.

\bibitem[Prajwal et~al.(2020)Prajwal, Mukhopadhyay, Namboodiri, and Jawahar]{10.1145/3394171.3413532}
K~R Prajwal, Rudrabha Mukhopadhyay, Vinay~P. Namboodiri, and C.V. Jawahar.
\newblock A lip sync expert is all you need for speech to lip generation in the wild.
\newblock In \emph{Proceedings of the 28th ACM International Conference on Multimedia}, page 484–492, New York, NY, USA, 2020. Association for Computing Machinery.

\bibitem[Ren et~al.(2021)Ren, Liu, and Zhao]{10.5555/3540261.3541331}
Yi Ren, Jinglin Liu, and Zhou Zhao.
\newblock Portaspeech: portable and high-quality generative text-to-speech.
\newblock In \emph{Proceedings of the 35th International Conference on Neural Information Processing Systems}, Red Hook, NY, USA, 2021. Curran Associates Inc.

\bibitem[Shen et~al.(2022)Shen, Li, Zhu, Duan, Zhou, and Lu]{10.1007/978-3-031-19775-8_39}
Shuai Shen, Wanhua Li, Zheng Zhu, Yueqi Duan, Jie Zhou, and Jiwen Lu.
\newblock Learning dynamic facial radiance fields for few-shot talking head synthesis.
\newblock In \emph{Computer Vision -- ECCV 2022}, pages 666--682, Cham, 2022. Springer Nature Switzerland.

\bibitem[Shen et~al.(2023)Shen, Zhao, Meng, Li, Zhu, Zhou, and Lu]{shen2023difftalk}
Shuai Shen, Wenliang Zhao, Zibin Meng, Wanhua Li, Zheng Zhu, Jie Zhou, and Jiwen Lu.
\newblock Difftalk: Crafting diffusion models for generalized audio-driven portraits animation.
\newblock In \emph{CVPR}, 2023.

\bibitem[Sinha et~al.(2022)Sinha, Biswas, Yadav, and Bhowmick]{Sinha2022EmotionControllableGT}
Sanjana Sinha, S. Biswas, Ravindra Yadav, and B. Bhowmick.
\newblock Emotion-controllable generalized talking face generation.
\newblock In \emph{International Joint Conference on Artificial Intelligence}, 2022.

\bibitem[Song et~al.(2021)Song, Meng, and Ermon]{song2021denoising}
Jiaming Song, Chenlin Meng, and Stefano Ermon.
\newblock Denoising diffusion implicit models.
\newblock In \emph{International Conference on Learning Representations}, 2021.

\bibitem[Stypułkowski et~al.(2024)Stypułkowski, Vougioukas, He, Zięba, Petridis, and Pantic]{10484496}
Michał Stypułkowski, Konstantinos Vougioukas, Sen He, Maciej Zięba, Stavros Petridis, and Maja Pantic.
\newblock Diffused heads: Diffusion models beat gans on talking-face generation.
\newblock In \emph{2024 IEEE/CVF Winter Conference on Applications of Computer Vision (WACV)}, pages 5089--5098, 2024.

\bibitem[Sun et~al.(2024)Sun, Xuan, Liu, and Xiang]{FGEmoTalk}
Zhaoxu Sun, Yuze Xuan, Fang Liu, and Yang Xiang.
\newblock Fg-emotalk: Talking head video generation with fine-grained controllable facial expressions.
\newblock \emph{Proceedings of the AAAI Conference on Artificial Intelligence}, 38:\penalty0 5043--5051, 2024.

\bibitem[Suzhen~Wang(2021)]{wang2021audio2head}
Yu~Ding Changjie Fan Xin~Yu Suzhen~Wang, Lincheng~Li.
\newblock Audio2head: Audio-driven one-shot talking-head generation with natural head motion.
\newblock In \emph{the 30th International Joint Conference on Artificial Intelligence (IJCAI-21)}, 2021.

\bibitem[Tan et~al.(2023)Tan, Ji, and Pan]{10378627}
Shuai Tan, Bin Ji, and Ye Pan.
\newblock Emmn: Emotional motion memory network for audio-driven emotional talking face generation.
\newblock In \emph{2023 IEEE/CVF International Conference on Computer Vision (ICCV)}, pages 22089--22099, 2023.

\bibitem[Tan et~al.(2024{\natexlab{a}})Tan, Ji, and Pan]{10658061}
Shuai Tan, Bin Ji, and Ye Pan.
\newblock Flowvqtalker: High-quality emotional talking face generation through normalizing flow and quantization.
\newblock In \emph{2024 IEEE/CVF Conference on Computer Vision and Pattern Recognition (CVPR)}, pages 26307--26317, 2024{\natexlab{a}}.

\bibitem[Tan et~al.(2024{\natexlab{b}})Tan, Ji, and Pan]{tan2024style2talker}
Shuai Tan, Bin Ji, and Ye Pan.
\newblock Style2talker: High-resolution talking head generation with emotion style and art style.
\newblock In \emph{Proceedings of the AAAI Conference on Artificial Intelligence}, pages 5079--5087, 2024{\natexlab{b}}.

\bibitem[Tan et~al.(2025)Tan, Ji, Bi, and Pan]{tan2025edtalk}
Shuai Tan, Bin Ji, Mengxiao Bi, and Ye Pan.
\newblock Edtalk: Efficient disentanglement for emotional talking head synthesis.
\newblock In \emph{European Conference on Computer Vision}, pages 398--416. Springer, 2025.

\bibitem[Tian et~al.(2025)Tian, Wang, Zhang, and Bo]{10.1007/978-3-031-73010-8_15}
Linrui Tian, Qi Wang, Bang Zhang, and Liefeng Bo.
\newblock Emo: Emote portrait alive generating expressive portrait videos with audio2video diffusion model under weak conditions.
\newblock In \emph{Computer Vision -- ECCV 2024}, pages 244--260, Cham, 2025. Springer Nature Switzerland.

\bibitem[Wang et~al.(2023{\natexlab{a}})Wang, Deng, Yin, Shum, and Wang]{wang2022pdfgc}
Duomin Wang, Yu Deng, Zixin Yin, Heung-Yeung Shum, and Baoyuan Wang.
\newblock Progressive disentangled representation learning for fine-grained controllable talking head synthesis.
\newblock In \emph{Proceedings of the IEEE/CVF Conference on Computer Vision and Pattern Recognition (CVPR)}, 2023{\natexlab{a}}.

\bibitem[Wang et~al.(2025)Wang, Weng, Li, Guo, Du, Niu, Ma, He, Wu, Hu, Yin, Liu, and Liu]{wang2025emotivetalk}
Haotian Wang, Yuzhe Weng, Yueyan Li, Zilu Guo, Jun Du, Shutong Niu, Jiefeng Ma, Shan He, Xiaoyan Wu, Qiming Hu, Bing Yin, Cong Liu, and Qingfeng Liu.
\newblock Emotivetalk: Expressive talking head generation through audio information decoupling and emotional video diffusion.
\newblock In \emph{Proceedings of the IEEE/CVF Conference on Computer Vision and Pattern Recognition (CVPR)}, 2025.
\newblock Poster Presentation \#1.

\bibitem[Wang et~al.(2023{\natexlab{b}})Wang, Zhao, Liu, Xu, Li, and Li]{inproceedings}
Jianrong Wang, Yaxin Zhao, Li Liu, Tianyi Xu, Qi Li, and Sen Li.
\newblock Emotional talking head generation based on memory-sharing and attention-augmented networks.
\newblock pages 2--6, 2023{\natexlab{b}}.

\bibitem[Wang et~al.(2004)Wang, Bovik, Sheikh, and Simoncelli]{1284395}
Zhou Wang, A.C. Bovik, H.R. Sheikh, and E.P. Simoncelli.
\newblock Image quality assessment: from error visibility to structural similarity.
\newblock \emph{IEEE Transactions on Image Processing}, 13\penalty0 (4):\penalty0 600--612, 2004.

\bibitem[Wei et~al.(2024)Wei, Yang, and Wang]{wei2024aniportrait}
Huawei Wei, Zejun Yang, and Zhisheng Wang.
\newblock Aniportrait: Audio-driven synthesis of photorealistic portrait animations, 2024.

\bibitem[Xu et~al.(2024)Xu, Li, Su, Shang, Zhang, Liu, Wang, Yao, and zhu]{xu2024hallo}
Mingwang Xu, Hui Li, Qingkun Su, Hanlin Shang, Liwei Zhang, Ce Liu, Jingdong Wang, Yao Yao, and Siyu zhu.
\newblock Hallo: Hierarchical audio-driven visual synthesis for portrait image animation, 2024.

\bibitem[Ye et~al.(2023)Ye, Jiang, Ren, Liu, He, and Zhao]{ye2023geneface}
Zhenhui Ye, Ziyue Jiang, Yi Ren, Jinglin Liu, Jinzheng He, and Zhou Zhao.
\newblock Geneface: Generalized and high-fidelity audio-driven 3d talking face synthesis.
\newblock In \emph{The Eleventh International Conference on Learning Representations}, 2023.

\bibitem[Zhang et~al.(2023)Zhang, Cun, Wang, Zhang, Shen, Guo, Shan, and Wang]{10204743}
Wenxuan Zhang, Xiaodong Cun, Xuan Wang, Yong Zhang, Xi Shen, Yu Guo, Ying Shan, and Fei Wang.
\newblock Sadtalker: Learning realistic 3d motion coefficients for stylized audio-driven single image talking face animation.
\newblock In \emph{2023 IEEE/CVF Conference on Computer Vision and Pattern Recognition (CVPR)}, pages 8652--8661, 2023.

\bibitem[Zhang et~al.(2021)Zhang, Li, Ding, and Fan]{zhang2021flow}
Zhimeng Zhang, Lincheng Li, Yu Ding, and Changjie Fan.
\newblock Flow-guided one-shot talking face generation with a high-resolution audio-visual dataset.
\newblock In \emph{Proceedings of the IEEE/CVF Conference on Computer Vision and Pattern Recognition}, pages 3661--3670, 2021.

\bibitem[Zheng et~al.(2024)Zheng, Zhang, Guo, Pan, Tan, Lu, Tang, An, and Yan]{zheng2024memo}
Longtao Zheng, Yifan Zhang, Hanzhong Guo, Jiachun Pan, Zhenxiong Tan, Jiahao Lu, Chuanxin Tang, Bo An, and Shuicheng Yan.
\newblock Memo: Memory-guided diffusion for expressive talking video generation.
\newblock \emph{arXiv preprint arXiv:2412.04448}, 2024.

\bibitem[Zhou et~al.(2021)Zhou, Sun, Wu, Loy, Wang, and Liu]{zhou2021pose}
Hang Zhou, Yasheng Sun, Wayne Wu, Chen~Change Loy, Xiaogang Wang, and Ziwei Liu.
\newblock Pose-controllable talking face generation by implicitly modularized audio-visual representation.
\newblock In \emph{Proceedings of the IEEE Conference on Computer Vision and Pattern Recognition (CVPR)}, 2021.

\end{thebibliography}
}

\end{document}